\documentclass{ecai} 
\usepackage{latexsym}
\usepackage{amssymb}
\usepackage{amsmath}
\usepackage{amsthm}
\usepackage{booktabs}
\usepackage{enumitem}
\usepackage{graphicx}
\usepackage{color}
\usepackage{algorithm}
\usepackage[noend]{algpseudocode}
\usepackage{multirow}
\usepackage{balance}
\usepackage{multicol}
\newcommand{\BibTeX}{B\kern-.05em{\sc i\kern-.025em b}\kern-.08em\TeX}

\begin{document}

%%%%%%%%%%%%%%%%%%%%%%%%%%%%%%%%%%%%%%%%%%%%%%%%%%%%%%%%%%%%%%%%%%%%%%%%

\begin{frontmatter}

%%% Use this command to specify your submission number.
%%% In doubleblind mode, it will be printed on the first page.

\paperid{123} 

%%% Use this command to specify the title of your paper.

\title{From Models to Network Topologies: A Topology Inference Attack in Decentralized Federated Learning}

%%% Use this combinations of commands to specify all authors of your 
%%% paper. Use \fnms{} and \snm{} to indicate everyone's first names 
%%% and surname. This will help the publisher with indexing the 
%%% proceedings. Please use a reasonable approximation in case your 
%%% name does not neatly split into "first names" and "surname".
%%% Specifying your ORCID digital identifier is optional. 
%%% Use the \thanks{} command to indicate one or more corresponding 
%%% authors and their email address(es). If so desired, you can specify
%%% author contributions using the \footnote{} command.

% \author[A]{\fnms{Chao}~\snm{Feng}\orcid{0000-0002-0672-1090}\thanks{Corresponding Author. Email: cfeng@ifi.uzh.ch.}}
% \author[A,B]{\fnms{Alberto}~\snm{Huertas Celdrán}\orcid{0000-0001-7125-1710}}
% \author[C]{\fnms{Gérôme}~\snm{Bovet}\orcid{0000-0002-4534-3483}} 
% \author[A]{\fnms{Burkhard}~\snm{Stiller}\orcid{0000-0002-7461-7463}} 

\author[A]{\fnms{Chao}~\snm{Feng}\thanks{Corresponding Author. Email: cfeng@ifi.uzh.ch.}}
\author[A]{\fnms{Yuanzhe}~\snm{Gao}}
\author[A,B]{\fnms{Alberto}~\snm{Huertas Celdrán}}
\author[C]{\fnms{Gérôme}~\snm{Bovet}} 
\author[A]{\fnms{Burkhard}~\snm{Stiller}} 
\address[A]{Communication Systems Group, Department of Informatics, University of Zurich, CH--8050 Zürich, Switzerland}
\address[B]{Department of Information and Communications Engineering, University of Murcia, 30100--Murcia, Spain}
\address[C]{Cyber-Defence Campus, armasuisse Science \& Technology, CH--3602 Thun, Switzerland}

%%% Use this environment to include an abstract of your paper.

\begin{abstract}
Federated Learning (FL) is widely recognized as a privacy-preserving Machine Learning paradigm due to its model-sharing mechanism that avoids direct data exchange. Nevertheless, model training leaves exploitable traces that can be used to infer sensitive information. In Decentralized FL (DFL), the topology, defining how participants are connected, plays a crucial role in shaping the model's privacy, robustness, and convergence. However, the topology introduces an unexplored vulnerability: attackers can exploit it to infer participant relationships and launch targeted attacks. This work uncovers the hidden risks of DFL topologies by proposing a novel Topology Inference Attack that infers the topology solely from model behavior. A taxonomy of topology inference attacks is introduced, categorizing them by the attacker's capabilities and knowledge. Practical attack strategies are designed for various scenarios, and experiments are conducted to identify key factors influencing attack success. The results demonstrate that analyzing only the model of each node can accurately infer the DFL topology, highlighting a critical privacy risk in DFL systems. These findings offer insights for improving privacy preservation in DFL environments.
\end{abstract}

\end{frontmatter}

%%%%%%%%%%%%%%%%%%%%%%%%%%%%%%%%%%%%%%%%%%%%%%%%%%%%%%%%%%%%%%%%%%%%%%%%

\section{Introduction}
\label{intro}
% intro of DFL
Federated Learning (FL) has emerged as a novel framework for enabling privacy-preserving Machine Learning (ML), facilitating collaborative model training among distributed clients without the necessity of sharing raw data~\cite{mcmahan2017communication}. Conventional FL systems are predicated on a centralized architecture, referred to as Centralized FL (CFL), wherein a central entity is responsible for collecting, aggregating, and redistributing models to the clients. Nevertheless, this centralized architecture presents several challenges, such as processing bottlenecks and the single points of failure risk~\cite{10251949}. In response to these limitations, Decentralized FL (DFL) has been introduced, wherein model training and aggregation occur locally, and communication between nodes relies on a peer-to-peer (P2P) network. This approach eliminates dependency on a central server, mitigating risks associated with single points of failure~\cite{beltran2024fedstellar}.

% information leakage in FL
Although FL, both centralized and decentralized, protect raw data privacy through its unique model-sharing mechanism, the training process inevitably leaves traces in the models. These traces can be exploited by malicious actors, potentially leading to inference attacks, including membership inference attacks, property inference attacks, and attribute inference attacks~\cite{hu2022membership}. Previous studies have illustrated that such attacks pose a considerable risk to privacy within FL systems by elucidating sensitive information based on the behavior of the models~\cite{dayal2023comparative}. While much of the existing work focuses on CFL, a notable research gap exists in exploring information leakage in DFL.

% topology
Overlay network topology defines how participants are interconnected in an FL system. In CFL, the client-server architecture enforces a fixed, star-shaped topology. However, DFL leverages the P2P network, enabling flexible node connections to form diverse topologies. Existing studies demonstrate that topology significantly influences DFL models' robustness and privacy-preserving capabilities~\cite{feng2024dart}. From a security perspective, exposure of the DFL network topology significantly facilitates malicious attacks. For example, in poisoning attack scenarios, topology awareness allows adversaries to pinpoint critical nodes, such as the central node in star topologies, thereby enabling them to amplify the poisoning intensity and enhance its propagation effectiveness~\cite{feng2024dart}. Similarly, for network-level attacks, knowledge of key nodes within the DFL topology can lead to more targeted and effective Distributed Denial-of-Service (DDoS) attacks. Thus, topology information must be regarded as a critical asset, both in terms of optimizing DFL training performance and ensuring system security. However, theoretical analyses on how network topology affects model convergence remain sparse. Moreover, there is limited research investigating potential information leakage risks for this sensitive overlay topology data and strategies to safeguard this critical asset.

%CONTRIBUTIONS
Therefore, this paper improves this research gap by proposing a novel overlay topology inference attack targeting DFL systems. The proposed attack leverages the behavioral traces generated by models to uncover sensitive overlay topologies. The main contributions of this work are as follows: (i) the formulation of a novel taxonomy of topology inference attacks, classifying these attacks based on the attacker's capabilities and knowledge; (ii) the development of practical strategies tailored to various attack types, accompanied by a quantitative analysis of critical factors that influence the efficiency of these attacks \footnote{Code available at: https://github.com/luke-feng/TIA}; (iii) rigorous experimental evaluations conducted across a range of datasets and real-world network topology configurations to empirically validate the proposed attack strategies; and (iv) the provision of insights aimed at informing the design of effective defensive mechanisms to safeguard sensitive information within DFL systems.

\section{Background and Related Work}
\label{rw}
This section overviews inference attacks, encompassing their classifications and potential attack surfaces. Since there is a lack of research directly related to topology inference attacks on DFL, this paper examines topology inference studies conducted in other domains, including communication systems and social networks.

\subsection{Inference Attacks}
Inference attacks exploit ML models to extract sensitive information without direct access to the underlying data~\cite{salem2018ml}. By analyzing model behaviors or outputs, adversaries can infer properties of the data, model parameters, or system structure. As FL grows in adoption for its privacy-preserving capabilities, it also faces increased vulnerability to such attacks~\cite{yin2021comprehensive}. Inference attacks vary in objectives and strategies, with common types including: 
\begin{itemize}
    \item \textbf{Membership Inference Attack:} Determines whether a specific data sample was part of the training dataset, often using shadow models or prediction confidence scores to differentiate between training and unseen data~\cite{7958568}. This is critical in sensitive domains like healthcare or finance.
    \item \textbf{Model Inversion Attack:} Reconstructs input data or infers its properties by iteratively optimizing inputs to match observed model outputs, posing risks in applications involving sensitive information like facial recognition or medical data~\cite{fredrikson2015model}.
    \item \textbf{Property Inference Attack:} Infers characteristics of the training data, such as demographic distributions, by analyzing model updates or outputs. In FL, this can involve monitoring node-specific behaviors to deduce shared properties of local datasets~\cite{ganju2018property}.  
    \item \textbf{Attribute Inference Attack:} Targets specific attributes of training data, often by injecting malicious samples to influence the model's learning process and make it vulnerable to data leakage~\cite{gong2018attribute}.    
\end{itemize}
These attacks not only compromise users' sensitive information and undermine data security but also erode trust in FL's privacy preserving mechanisms. Users may hesitate to participate in the training process, undermining the overall effectiveness of the FL system. However, existing research primarily focuses on CFL, with limited exploration of privacy leakage in DFL. Moreover, these studies have not adequately addressed the critical role of DFL's overlay topology.

\subsection{Topology Inference}
Topology inference is explored in various research domains, such as identifying source–destination paths in communication systems and uncovering interconnections among nodes in social networks.

\subsubsection{Communication Systems}
In communication systems, most protocols are designed to avoid ring formation, resulting in tree-like topologies. Many topology inference studies commonly assume a single path between nodes~\cite{castro2004network}. A widely used approach involves leveraging ICMP-based protocols, such as the traceroute, to map network topologies~\cite{jin2006network}. Tomography-based metrics, like packet loss rates, further aid in deducing a system's topology: by observing loss rates from a source node to a target node, one can estimate the likelihood of a direct connection and thereby reconstruct the network~\cite{coates2002internet}. Additionally, by measuring the timestamps of sent and received packets, it is possible to determine the number of forwarding hops, offering deeper insight into the overall network configuration~\cite{hou2020proto}.

\subsubsection{Social Networks}
Social networks generally refer to the social structure formed by individuals in society connected by a specific relationship. In social networks, user interactions (e.g., likes, retweets) serve as signals for topology modeling and inference by capturing how information diffuses across the network~\cite{gomez2012inferring}. However, this behavioral data is often incomplete, especially in large-scale settings, necessitating methods like noisy sparse subspace clustering to infer network structures from partial observations~\cite{wang2016graph}. Meanwhile, social networks are dynamic and time-sensitive, and they frequently exhibit small-world properties.\cite{gong2012evolution} incorporate these distributional patterns as prior knowledge in evolutionary models to better capture network changes over time.

To conclude, communication networks often assume tree topologies and social networks are typically modeled as small-world networks. These assumptions do not hold for DFL network topologies. Moreover, common inference tools and metrics, such as traceroute, packet loss rate, or user interaction data, are not readily applicable in DFL contexts, creating the need for specialized inference metrics and strategies tailored to the particularities of DFL systems.

\section{Overlay Topology in DFL}
\label{overlay}
DFL is an ML paradigm that facilitates collaborative model training across multiple nodes without a central server. In DFL, each node updates its local model parameters using its private data and exchanges them with its directly connected neighbors. As the overlay topology in DFL significantly affects model convergence, this work presents the topology as an undirected graph \( G = (V, E) \), where \( V \) is the set of nodes and \( E \) represents the edges indicating direct communication links between nodes. The adjacency matrix \( A \) encodes the graph structure, where:
\begin{equation}
A_{ij} = 
\begin{cases}
1 & \text{if nodes \( i \) and \( j \) are connected}, \\
0 & \text{otherwise}.
\end{cases}
\label{eq:adj}
\end{equation}
The degree matrix \( D \) is a diagonal matrix where \( D_{ii} = \text{degree}(i) + 1 \). The aggregation process is governed by the normalized aggregation matrix:

\begin{equation}
    P = D^{-1}(A + I),
    \label{eq:agg_mat}
\end{equation}

where \( I \) is the identity matrix. This formulation ensures \( P \) is row-stochastic (\(\sum_j P_{ij} = 1\)), enabling consistent scaling during aggregation.

DFL proceeds iteratively over \( T \) rounds, comprising local training and model aggregation steps. Let \( M_t = [\theta_1^{(t)}, \theta_2^{(t)}, \dots, \theta_N^{(t)}] \) represent the models of all nodes at round \( t \), where \( \theta_i^{(t)} \) denotes the parameters of node \( i \).

During the training stage, each node updates its model using its local dataset:

\begin{equation}
M_t = \tilde{M}_{t-1} + \delta_t,
\end{equation}

where \( \delta_t \) represents local updates, and \( \tilde{M}_{t-1} \) is the aggregated model from round \( t-1 \).

In the aggregation stage, nodes aggregate models from neighbors using the aggregation matrix \( P \):

\begin{equation}
\tilde{M}_t = P M_{t-1}.
\end{equation}

The model parameters after \( T \) rounds can be expressed as:
\begin{equation}
M_T = P^T M_0 + \sum_{t=1}^T P^{T-t} \delta_t.
\label{eq:dfl_learning}
\end{equation}

Assuming \( G \) is connected (i.e., no isolated nodes), the spectral radius of \( P \) satisfies \( \rho(P) < 1 \). Additionally, under bounded local updates (e.g., using SGD or Adam), there exists \( C > 0 \) such that \( \|\delta_i^{(t)}\| \leq C \) for all \( t \) and \( i \). As \( T \) increases, the influence of the initial model diminishes:

\begin{equation}
\|P^T M_0\| \leq \|P^T\| \|M_0\| \rightarrow 0 \quad \text{as} \quad T \rightarrow \infty.
\label{eq:importance}
\end{equation}

The cumulative effect of local updates is:

\begin{equation}
\sum_{t=1}^T P^{T-t} \delta_t,
\end{equation}

where \( \|P^{T-t} \delta_t\| \) is bounded by \( \|P^{T-t}\| \|\delta_t\| \). Since \( \|P^{T-t}\| \rightarrow 0 \) as \( T \rightarrow \infty \), the impact of individual updates diminishes, and the series converges due to \( \rho(P) < 1 \).

Thus, it can be seen that the convergence behavior of DFL is influenced by the topology captured by \( P \). Initially, the aggregation dynamics (powers of \( P \)) dominate, reducing the influence of \( M_0 \). Over time, local updates drive the learning process, modulated by the decaying powers of \( P \). This interplay underscores the importance of overlay topology in determining convergence efficiency.

Given the critical role of overlay topology, it is essential to investigate potential information leakage through model updates and develop effective methods to protect sensitive topology information in DFL.

\section{Problem Statement}

Theoretical analysis of the DFL learning process reveals that overlay topology is crucial in determining system performance, robustness, and privacy. This section introduces the problem of topology inference and categorizes potential attack types based on the attacker's knowledge and capabilities.

Consider a DFL system represented by an undirected graph \( G = (V, E) \), where \( V \) is the set of participating nodes, and \( E \) is the set of edges representing direct communication links between nodes. The adjacency matrix \( A \) of size \( |V| \times |V| \) encodes the network topology, as defined in Equation \eqref{eq:adj}.

An attacker aims to infer a predicted adjacency matrix \( A' \) that approximates the ground truth \( A \) as closely as possible, thereby revealing sensitive topology information.

The effectiveness of topology inference attacks depends on the attacker's knowledge and capabilities. The following assumptions are made. \textbf{(i) Internal Adversary}: The attacker is an internal participant in the DFL system and can identify a subset of nodes. \textbf{(ii) Decoupled Information}: Node-level information (e.g., models and datasets) and network-level information (e.g., connections) are treated as separate assets. For instance, an attacker may know a node's model but not its connections.

Let \( V' \subset V \) represent a set of nodes known to the attacker and \( E' \subset E \) represent the subset of edges known to the attacker, where \( \emptyset \subsetneq E' \subsetneq E \) to represent partial knowledge. The attacker's goal is to infer the entire edge set \( E \) based on partial knowledge of \( V' \) and \( E' \).

For each node \( i \in V \), the local model is represented as \( M_i \), and the local dataset is represented as \( D_i \). This paper assumes that models and data within a node are distinct assets. Thus, the attacker's capabilities are defined as follows:

\begin{itemize}
    \item \textbf{Model Knowledge}: The attacker can access models of known nodes \( V' \), forming the set:  \( M' = \{ M_i \mid i \in V' \}. \)
    \item \textbf{Dataset Knowledge}: The attacker may not have access to the datasets \( D_i \) of these nodes. The datasets known to the attacker are denoted by: \( D' \subseteq \{ D_i \mid i \in V' \} \quad \text{or} \quad D' = \emptyset.\)
\end{itemize}

Based on the attacker's knowledge and capabilities, the topology inference attack is classified into five scenarios: 
\begin{itemize}

    \item \textbf{Scenario 1:} The attacker knows all node models and datasets, but only partial edge information (\( M' = M, D' = D,  \emptyset \subsetneq E' \subsetneq E \)).
    \item \textbf{Scenario 2:} The attacker knows all node models, but none of the datasets, and only partial edge information (\( M' = M, D' = \emptyset,  \emptyset \subsetneq E' \subsetneq E \)).
    \item \textbf{Scenario 3:} The attacker knows all node models and datasets, but does not have edge information (\( M' = M, D' = D,  E' = \emptyset \)).
    \item \textbf{Scenario 4:} The attacker knows all node models, but neither datasets nor edge information (\( M' = M, D' = \emptyset,  E' = \emptyset \)).  
    \item \textbf{Scenario 5:} The attacker knows partial models, but neither datasets nor edge information (\( \emptyset \subsetneq M' \subsetneq M, D' = \emptyset, E' = \emptyset \)).
\end{itemize}

For Scenario 1, since the attacker can control all nodes and knows partial information about the edges, it can be classified as a white-box attack. Scenarios 2, 3, and 4 can be categorized as gray-box attacks. For attack scenario 5, no information is available regarding the total number of nodes or edges, making it difficult to fully reconstruct the network topology. %Thus, this work mainly considers the attack scenarios 1, 2, 3, and 4.

\section{Topology Inference Attack}
This section first explores the metrics that attackers can exploit to carry out topology inference attacks, and then designs attack strategies for various attack scenarios.
\subsection{Attack Metrics}
In a DFL system, attackers may leverage the local data and local models available at each known node. Additionally, DFL's decentralized communication mechanism allows attackers to access models transmitted from neighboring nodes. Thus, this paper proposes six attack metrics that attackers can exploit:

\begin{itemize}

    \item \textbf{Relative Loss.} This metric evaluates how well a model trained on one node's data generalizes to another node's dataset. Consider a model \( f_i \) trained on node \( i \)'s dataset \( D_i \), and let \( D_j \) be the dataset of another node \( j \). The relative loss \(\text{Relative Loss}_{i,j}\) measures the performance of \( f_i \) on \( D_j \). A lower relative loss indicates better transferability and generalization across nodes:
    \begin{equation}
    \text{Relative Loss}_{i,j} = \mathcal{L}(f_i, D_j)
    \label{eq:loss}
    \end{equation}
    where \(\mathcal{L}(f, D)\) is the loss function used by model \(f\) on dataset \(D\).

    \item \textbf{Relative Entropy.} This metric measures the uncertainty of \( f_i \)'s predictions on \( D_j \). While relative loss focuses on correctness, relative entropy focuses on confidence. A model may be confident (low entropy) but still perform poorly if it is incorrect.
    \begin{equation}
    \text{Relative Entropy}_{i,j} = - \frac{1}{|D_j|} \sum_{x \in D_j} \sum_{k} y_k(x) \log(f_{i,k}(x))
    \label{eq:entropy}
    \end{equation}
    where \(y_k(x)\) is the true label distribution for sample \(x\) and \(f_{i,k}(x)\) is the predicted probability of class \(k\) under model \(f_i\).

    \item \textbf{Relative Sensitivity.} This metric quantifies the sensitivity of model \( f_i \)'s predictions on dataset \( D_j \) using the Jacobian norm. Relative sensitivity could capture the model's susceptibility to small input perturbation.

    \begin{equation}
    \text{Relative Sensitivity}_{i,j} = \frac{1}{|D_j|}\sum_{x \in D_j}\left\|\frac{\partial f_i(x)}{\partial x}\right\|_F
    \label{eq:sensitivity}
    \end{equation}
    
    where \(\frac{\partial f_i(x)}{\partial x}\) is the Jacobian of \(f_i\)'s outputs with respect to input \(x\), and \(\|\cdot\|_F\) denotes the Frobenius norm.

    % \item \textbf{Relative Curvature.} This metric quantifies the curvature change of model updates between consecutive training rounds. Unlike previous metrics that only consider single-round scenarios, relative curvature captures temporal dynamics by computing the Hessian norm of differences between models \( f_i^{t} \) and \( f_i^{t-1} \) on dataset \( D_j \). Higher curvature indicates increased sensitivity to model updates, suggesting potential training instability.

    % \begin{equation}
    % \text{Relative Curvature}_{i,j}^{t} = \frac{1}{|D_j|}\sum_{x \in D_j}\left\|\nabla_{x}^2 \left(f_i^{t}(x)-f_i^{t-1}(x)\right)\right\|_F
    % \label{eq:curvature}
    % \end{equation}
    
    % where \(\nabla_{x}^2(\cdot)\) denotes the Hessian operator with respect to input \( x \).
    
    \item \textbf{Cosine Similarity.} This metric compares the direction of two parameter vectors \(\mathbf{a}\) and \(\mathbf{b}\) derived from different models. If both vectors point in similar directions, their cosine similarity is high. In DFL, a high cosine similarity suggests that the two nodes frequently aggregate their parameters, resulting in more closely aligned models:
    \begin{equation}
    \text{Cosine Similarity}(\mathbf{a}, \mathbf{b}) = \frac{\mathbf{a} \cdot \mathbf{b}}{\|\mathbf{a}\| \, \|\mathbf{b}\|}
    \label{eq:cosine_similarity}
    \end{equation}
    where \(\mathbf{a} \cdot \mathbf{b}\) is the dot product, and \(\|\mathbf{a}\|\), \(\|\mathbf{b}\|\) are the L2 norms of the vectors \(\mathbf{a}\) and \(\mathbf{b}\), respectively.

    \item \textbf{Euclidean Similarity.} It quantifies how similar two vectors are based on the Euclidean distance between them.  A higher Euclidean similarity indicates that the two models are numerically closer, while a smaller distance suggests they differ significantly:
    \begin{equation}
    \text{Euclidean Similarity}(\mathbf{a}, \mathbf{b}) = \frac{1}{1+\sqrt{\sum_{i=1}^{n} (a_i - b_i)^2}}
    \label{eq:euclidean_similarity}
    \end{equation}
    where \(a_i\) and \(b_i\) are the \(i\)-th elements of \(\mathbf{a}\) and \(\mathbf{b}\).   
   
    \item \textbf{Curvature Divergence.} This metric quantifies model divergence by analyzing implicit second-order information from consecutive parameter updates, reflecting curvature characteristics without explicit data dependence. Specifically, it compares parameter update vectors to capture similarity in model evolution indirectly:

    \begin{equation}
    \text{Curvature Divergence}(\Delta \mathbf{a}, \Delta \mathbf{b}) 
    = \frac{\|\Delta \mathbf{a}-\Delta \mathbf{b}\|_2}{\frac{1}{2}(\|\Delta \mathbf{a}\|_2+\|\Delta \mathbf{b}\|_2)}
    \label{eq:curvature_similarity}
    \end{equation}
    
    where \(\Delta \mathbf{a} = \mathbf{a}^{t}-\mathbf{a}^{t-1}\) and \(\Delta \mathbf{b} = \mathbf{b}^{t}-\mathbf{b}^{t-1}\) denote parameter updates between rounds \(t-1\) and \(t\). A lower value indicates lower divergence (i.e., more similar updates).
\end{itemize}

Relative metrics, including relative loss, entropy, and sensitivity, require access to both local models and local datasets, thus making them suitable for \textbf{attack scenarios 1} and \textbf{3}. While similarity metrics, \textit{i.e.} cosine, Euclidean, and curvature similarity, depend only on the models and do not require dataset access, they are appropriate for \textbf{attack scenarios 2} and \textbf{4}. %The detailed justification of curvature metrics for capturing model similarity is provided in the supplementary material.

\subsection{Attack Strategies}
In a topology inference attack, the attacker's objective is to construct a predicted adjacency matrix \( A' \) that closely approximates the ground-truth adjacency matrix  \( A \). 
% This can be viewed as a binary classification problem, determining whether each element  \( A'_{ij} \) in \( A' \) equals one or zero. 

\subsubsection{Supervised Attack Strategy.} 
In \textbf{attack scenarios 1} and \textbf{2}, where the attacker has access to partial ground-truth edge labels, this work introduces \textit{EDGEPRE}, a supervised edge classification framework for topology reconstruction. As shown in Algorithm~\ref{alg:edgepre}, the model learns to infer the full adjacency matrix from node-level behavioral features and a small set of labeled node pairs.

\paragraph{Model Architecture.}
Given $N$ nodes and their behavioral metrics \( X \in \mathbb{R}^{N \times d} \), \textit{EDGEPRE} directly uses an MLP-based edge classifier without any message passing. For each labeled pair \( (i, j) \), the concatenated feature pair \( [x_i \| x_j] \) is passed to a multilayer perceptron (MLP) decoder that outputs the predicted edge probability:
\[
\hat{y}_{ij} = \sigma(f_{\text{dec}}([x_i \| x_j])).
\]
Optionally, feature interaction terms such as \( x_i * x_j \) and \( |x_i - x_j| \) can be included to enhance the representation.

\paragraph{Training Objective.}
The model is optimized using binary cross-entropy over the labeled edge set:
\[
\mathcal{L} = \sum_{(i,j,y_{ij}) \in \mathcal{P}} \text{BCE}(\hat{y}_{ij}, y_{ij}).
\]

\paragraph{Adjacency Reconstruction.}
After training, \textit{EDGEPRE} generalizes to all node pairs and produces a reconstructed soft adjacency matrix \( A' \in [0,1]^{N \times N} \), where each entry is computed via:
\[
A'_{ij} = \sigma(f_{\text{dec}}([x_i \| x_j])).
\]

\begin{algorithm}
\caption{\textsc{EDGEPRE (MLP-based)}}
\label{alg:edgepre}
\begin{algorithmic}[1]
\Require Node features $X \in \mathbb{R}^{N \times d}$, labeled edge set $\mathcal{P}_{\text{train}} = \{(i, j, y_{ij})\}$, number of epochs $T$
\Ensure Reconstructed adjacency matrix $A' \in [0, 1]^{N \times N}$

\State Initialize MLP decoder $f_{\text{dec}}$
\For{$t = 1$ to $T$}
    \For{each labeled pair $(i, j, y_{ij}) \in \mathcal{P}_{\text{train}}$}
        \State $h_{ij} \gets [x_i \| x_j]$
        \State $\hat{y}_{ij} \gets \sigma(f_{\text{dec}}(h_{ij}))$
    \EndFor
    \State Compute loss: $\mathcal{L} \gets \sum_{(i,j)} \text{BCE}(\hat{y}_{ij}, y_{ij})$
    \State Update decoder parameters via gradient descent
\EndFor
\vspace{0.3em}
\State \textit{\textbf{Inference phase}}
\ForAll{$(i, j)$ where $1 \leq i, j \leq N$}
    \State $h_{ij} \gets [x_i \| x_j]$
    \State $A'_{ij} \gets \sigma(f_{\text{dec}}(h_{ij}))$
\EndFor
\State \Return $A'$
\end{algorithmic}
\end{algorithm}

\subsubsection{Unsupervised Attack Strategy.}
For \textbf{attack scenarios 3} and \textbf{4}, where no edge labels are available, this work proposes \textit{INFERGAT}, an unsupervised structure inference method based on a GAT encoder-decoder framework. The model predicts the latent adjacency purely from node-level behavioral metrics without requiring ground-truth edges.

\paragraph{Model Architecture.}
The input \( X \in \mathbb{R}^{N \times d} \) contains attack-related features such as similarity or entropy. A GAT encoder outputs embeddings \( Z = f_{\text{enc}}(X) \), and a symmetric MLP decoder estimates pairwise edge strength via \( A'_{ij} = \sigma(\text{MLP}([z_i \| z_j])) \).

\paragraph{Training Objective.}
The model is trained to reconstruct the behavioral structure with mean squared error:
\begin{equation}
\mathcal{L} = \sum_{i,j} \left\| A'_{ij} - X_{ij} \right\|^2
\end{equation}
The final output is the predicted adjacency matrix \( A' \in [0,1]^{N \times N} \).

\begin{algorithm}[t]
\caption{\textsc{INFERGAT}}
\label{alg:gat-infer}
\begin{algorithmic}[1]
\Require Attack metrics $X \in \mathbb{R}^{N \times d}$, learning rate $\eta$, epochs $T$
\Ensure Predicted adjacency matrix $A' \in [0, 1]^{N \times N}$

\State Initialize GAT encoder $f_{\text{enc}}$ and decoder $f_{\text{dec}}$ with parameters $\theta$
\For{$t = 1$ to $T$}
    \State $Z \gets f_{\text{enc}}(X)$ \Comment{Node embeddings via multi-head GAT}
    \State $A'_{ij} \gets f_{\text{dec}}(z_i, z_j),\quad \forall i,j$ \Comment{Decoder predicts edge weights}
    \State $A' \gets \frac{1}{2}(A' + A'^\top)$ \Comment{Symmetrize adjacency}
    \State $\mathcal{L} \gets \sum_{i,j} \left\| A'_{ij} - X_{ij} \right\|^2$ \Comment{Reconstruction loss}
    \State $\theta \gets \theta - \eta \cdot \nabla_\theta \mathcal{L}$ \Comment{Gradient descent update}
\EndFor
\State \Return $A'$
\end{algorithmic}
\label{gat-infer}
\end{algorithm}

\textit{EDGEPRE} and \textit{INFERGAT} represent two attack strategies for topology inference in DFL. \textit{EDGEPRE} leverages partial edge supervision to perform explicit edge classification, while \textit{INFERGAT} relies solely on behavioral metrics to reconstruct structural patterns in a self-supervised manner. Together, they cover a broad range of adversarial capabilities and reflect realistic threat models across different attack scenarios.

\section{Evaluation}
This work empirically validates the effectiveness of the proposed topology inference attack. First, it identifies which metrics yield stronger attack performance. Building on these results, it then thoroughly evaluates the proposed attack strategies under various conditions, including different datasets, topologies, node counts, and data preprocessing methods.

\subsection{Experimental Setups}
This section describes the configurations employed in the experiments.

\textbf{A. Dataset and Model:} Experiments are conducted on: MNIST, FMNIST, CIFAR10, SVHN, ImageNet10, and Malware, which are widely used in FL and inference attack benchmarks.

\begin{itemize}
\item \textbf{MNIST \cite{lecun2010mnist}}: Contains 60,000 grayscale training images and 10,000 test images of handwritten digits (0–9) at 28$\times$28 resolution. A two-hidden-layer MLP (256 and 128 neurons) trained with Adam (lr=1e-3) is used.

\item \textbf{FMNIST \cite{xiao2017fashion}}: Similar size and format to MNIST, but with 10 categories of clothing items. A CNN with two convolutional layers (32 and 64 filters, kernel size 3$\times$3) is employed.

\item \textbf{CIFAR10 \cite{krizhevsky2009learning}}: Contains 60,000 32$\times$32 RGB images from 10 categories. The training set is only treated with basic image normalization. The image pixel values are normalized to keep the original image content unchanged. A MobileNet~\cite{sandler2019mobilenetv2invertedresidualslinear} is used for training.

\item \textbf{SVHN \cite{netzer2011reading}}: Composes of over 600,000 32$\times$32 RGB images of house numbers collected from Google Street View. Each image contains a centered digit from 0 to 9. A ResNet9~\cite{he2016deep} is adopted as the training model.

\item \textbf{ImageNet10}: A curated subset of the ImageNet~\cite{deng2009imagenet}, containing 10 classes selected from the original ImageNet hierarchy. This version~\cite{kaggle_imagenet100} includes approximately 15,000 images with varied object categories and balanced class distributions. A lightweight PoolFormer-S12~\cite{yu2022metaformer} model is used for training.

\item \textbf{Malware \cite{feng2025crowdsensing}}: The Malware dataset comprises 342,106 tabular records with 31 extracted features, collected from eight IoT crowdsensing devices infected by eight types of malware. Each record is labeled as either the corresponding malware type or normal behavior. A MLP with two hidden layers (64 and 32 neurons) is employed.
\end{itemize}

\textbf{B. DFL Topology Setting:} All experiments are conducted on DFL systems comprising two categories of topologies: synthetic and real-world. The synthetic topologies include ring, star, and three Erdős-Rényi (ER) graphs with connection probabilities \( p = \)0.3, 0.5, and 0.7. The real-world topologies consist of 27 network topologies collected from the SNDlib~\cite{orlowski2010sndlib} and DEFO~\cite{defo2015topology} repositories, which reflect realistic communication and infrastructure networks. Detailed descriptions and properties of these topologies are provided in the supplementary material.

\textbf{C. Federation Setting:} All experiments employ the following DFL configurations:
\begin{itemize}   
    \item \textbf{Number of Nodes:} For synthetic topologies, the number of nodes is set to 10, 20, and 30. For real-world topologies, the number of nodes ranges from 12 to 87, depending on the specific topology.
    \item \textbf{Total Rounds:} The total number of communication rounds is determined by the rule \textit{rounds = number of nodes}. The mechanism ensures efficient inter-node communication, thereby enabling consistent scalability of training time as the network size increases.
    \item \textbf{Local Epochs:} Two scenarios (3 and 10 local epochs) are considered to evaluate how varying levels of local overfitting affect the attack's effectiveness.
    \item \textbf{Data Distribution:} Both Independent and Identically Distributed (IID) and non-IID \( \alpha = 0.1 \) data distributions are considered to evaluate the impact of data heterogeneity.
\end{itemize}

\textbf{D. Attack Models:} To evaluate the effectiveness of the proposed methods, \textit{EDGEPRE} is compared against classical supervised classifiers, including Logistic Regression, Support Vector Machine (SVM), and Random Forest (RF). 
Additionally, graph-specific supervised baselines are also used, including: \textit{SEAL}~\cite{zhang2018seal} and \textit{GraphSAGE} (edge classification setting)~\cite{hamilton2017inductive}.

For the unsupervised attack strategy, \textit{INFERGAT} is evaluated against standard clustering-based methods, including K-Means, Gaussian Mixture Model (GMM), and Spectral Clustering. 
This experiment further adds modern graph-reconstruction and self-supervised baselines: \textit{GRACE}~\cite{zhu2020grace} and \textit{BGRL}~\cite{thakoor2021bgrl}.

\textbf{E. Evaluation Metric:} Since the topology inference attack is formulated as a binary classification problem, evaluation uses the \textbf{F1-Score} and \textbf{AUC-ROC} metrics.

\subsection{Selection of Attack Metrics}
\begin{figure}[t]
    \centering
    \includegraphics[width=1\columnwidth]{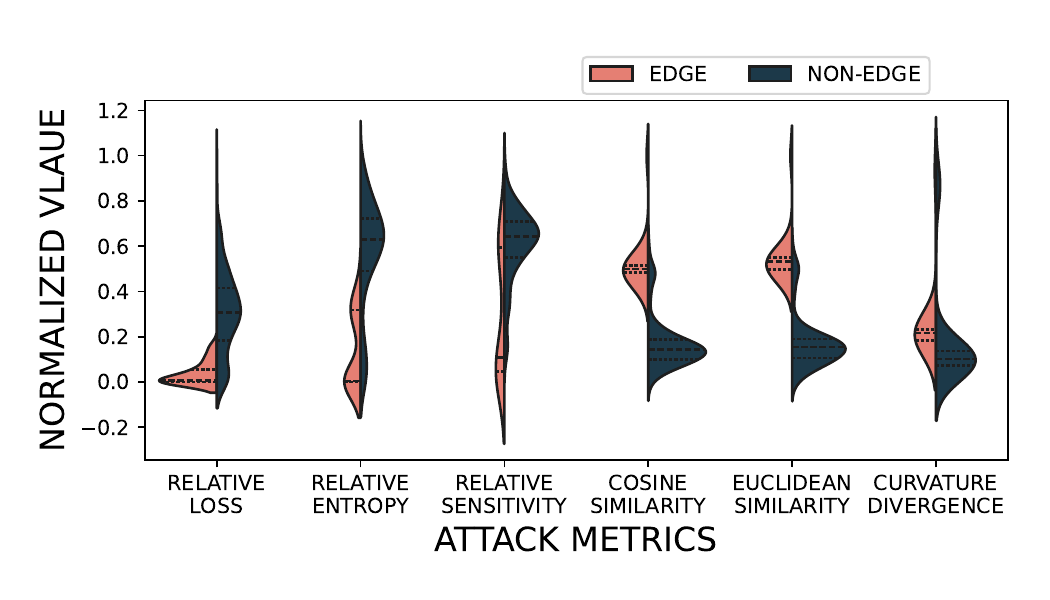}
    \caption{Normalized distributions of attack metrics in a 20-node star-topology DFL system trained on the FMNIST dataset. The “edge group” corresponds to metrics calculated between directly connected nodes, while the “non-edge group” corresponds to metrics calculated between nodes with no direct connection.}
    \label{fig:metrics_distibution}
\end{figure}
Selecting appropriate attack metrics is critical to the success of inference attacks. An effective attack metric should distinguish between connected and non-connected nodes as clearly as possible. To this end, the first part of this experiment examines the effectiveness of the six proposed attack metrics by analyzing their respective distributions. 

\figurename~\ref{fig:metrics_distibution} illustrates the distributions of these metrics for a 20-node star topology DFL system trained on the FMNIST dataset, comparing directly connected nodes (edge group) with those not (non-edge group). To facilitate direct comparison, values have been normalized. 

The results indicate that the relative loss for connected nodes is notably lower than that for non-connected nodes, with minimal overlap between the respective distributions. By contrast, the relative entropy and sensitivity metrics exhibit substantially greater overlap. Since computing the relative loss requires access to local models and data, it has been employed in attack scenarios 1 and 3.

\begin{table*}[t]
    \centering
    % \scriptsize
    \setlength{\tabcolsep}{2.5pt}
    \resizebox{\textwidth}{!}{
    \begin{tabular}{l l c c c c c c c c c c c c}
    \toprule Scenario & Algorithm  & \multicolumn{2}{c}{CIFAR10}  & \multicolumn{2}{c}{FMNIST}  & \multicolumn{2}{c}{ImageNet10}  & \multicolumn{2}{c}{Malware}  & \multicolumn{2}{c}{MNIST}  & \multicolumn{2}{c}{SVHN}  \\
 &  & F1 & AUC & F1 & AUC & F1 & AUC & F1 & AUC & F1 & AUC & F1 & AUC \\ \midrule
SC1 & Logistic & 0.806 $\pm$ 0.129 & 0.519 $\pm$ 0.064 & 0.804 $\pm$ 0.126 & 0.500 $\pm$ 0.000 & 0.804 $\pm$ 0.126 & 0.500 $\pm$ 0.000 & 0.804 $\pm$ 0.126 & 0.500 $\pm$ 0.000 & 0.800 $\pm$ 0.125 & 0.500 $\pm$ 0.000 & 0.804 $\pm$ 0.126 & 0.500 $\pm$ 0.000 \\
 & RF & 0.795 $\pm$ 0.132 & 0.643 $\pm$ 0.127 & 0.786 $\pm$ 0.142 & 0.653 $\pm$ 0.128 & 0.833 $\pm$ 0.107 & 0.695 $\pm$ 0.132 & 0.714 $\pm$ 0.144 & 0.503 $\pm$ 0.050 & 0.765 $\pm$ 0.135 & 0.600 $\pm$ 0.102 & 0.788 $\pm$ 0.114 & 0.633 $\pm$ 0.116 \\
 & SVM & 0.840 $\pm$ 0.102 & 0.609 $\pm$ 0.153 & 0.818 $\pm$ 0.115 & 0.561 $\pm$ 0.122 & 0.808 $\pm$ 0.118 & 0.519 $\pm$ 0.066 & 0.800 $\pm$ 0.137 & 0.499 $\pm$ 0.025 & 0.801 $\pm$ 0.131 & 0.519 $\pm$ 0.067 & 0.824 $\pm$ 0.108 & 0.566 $\pm$ 0.130 \\

 & GraphSage & 0.740 $\pm$ 0.077 & 0.607 $\pm$ 0.253 & 0.699 $\pm$ 0.058 & 0.474 $\pm$ 0.188 & 0.701 $\pm$ 0.045 & 0.507 $\pm$ 0.209 & 0.686 $\pm$ 0.043 & 0.423 $\pm$ 0.186 & 0.686 $\pm$ 0.039 & 0.460 $\pm$ 0.185 & 0.712 $\pm$ 0.056 & 0.526 $\pm$ 0.200 \\
 & SEAL & 0.701 $\pm$ 0.057 & 0.560 $\pm$ 0.142 & 0.697 $\pm$ 0.052 & 0.556 $\pm$ 0.152 & 0.699 $\pm$ 0.057 & 0.557 $\pm$ 0.156 & 0.705 $\pm$ 0.052 & 0.577 $\pm$ 0.144 & 0.701 $\pm$ 0.050 & 0.584 $\pm$ 0.129 & 0.697 $\pm$ 0.056 & 0.548 $\pm$ 0.146 \\
  & \textbf{EDGEPRE} & \textbf{0.847 $\pm$ 0.101} & \textbf{0.831 $\pm$ 0.136} & \textbf{0.873 $\pm$ 0.096} & \textbf{0.857 $\pm$ 0.135} & \textbf{0.855 $\pm$ 0.101} & \textbf{0.826 $\pm$ 0.181} & \textbf{0.858 $\pm$ 0.102} & \textbf{0.834 $\pm$ 0.192} & \textbf{0.862 $\pm$ 0.095} & \textbf{0.839 $\pm$ 0.144} & \textbf{0.886 $\pm$ 0.091} & \textbf{0.871 $\pm$ 0.130} \\
\midrule
SC2 & Logistic & 0.811 $\pm$ 0.128 & 0.543 $\pm$ 0.101 & 0.804 $\pm$ 0.126 & 0.500 $\pm$ 0.000 & 0.801 $\pm$ 0.127 & 0.500 $\pm$ 0.001 & 0.804 $\pm$ 0.126 & 0.500 $\pm$ 0.000 & 0.800 $\pm$ 0.125 & 0.500 $\pm$ 0.000 & 0.804 $\pm$ 0.126 & 0.500 $\pm$ 0.000 \\
 & RF & 0.787 $\pm$ 0.134 & 0.629 $\pm$ 0.132 & 0.762 $\pm$ 0.135 & 0.568 $\pm$ 0.107 & 0.784 $\pm$ 0.122 & 0.579 $\pm$ 0.096 & 0.722 $\pm$ 0.139 & 0.505 $\pm$ 0.048 & 0.746 $\pm$ 0.140 & 0.562 $\pm$ 0.082 & 0.765 $\pm$ 0.129 & 0.598 $\pm$ 0.102 \\
 & SVM & \textbf{0.836 $\pm$ 0.108} & 0.604 $\pm$ 0.151 & 0.817 $\pm$ 0.117 & 0.559 $\pm$ 0.119 & 0.810 $\pm$ 0.116 & 0.520 $\pm$ 0.067 & 0.800 $\pm$ 0.137 & 0.499 $\pm$ 0.025 & 0.800 $\pm$ 0.132 & 0.517 $\pm$ 0.066 & 0.821 $\pm$ 0.110 & 0.560 $\pm$ 0.128 \\
 
 & GraphSage & 0.697 $\pm$ 0.050 & 0.488 $\pm$ 0.172 & 0.700 $\pm$ 0.069 & 0.510 $\pm$ 0.152 & 0.699 $\pm$ 0.050 & 0.501 $\pm$ 0.135 & 0.700 $\pm$ 0.056 & 0.473 $\pm$ 0.157 & 0.698 $\pm$ 0.056 & 0.498 $\pm$ 0.139 & 0.699 $\pm$ 0.060 & 0.490 $\pm$ 0.152 \\
 & SEAL & 0.699 $\pm$ 0.042 & 0.558 $\pm$ 0.137 & 0.704 $\pm$ 0.052 & 0.582 $\pm$ 0.108 & 0.700 $\pm$ 0.046 & 0.589 $\pm$ 0.105 & 0.700 $\pm$ 0.050 & 0.561 $\pm$ 0.145 & 0.701 $\pm$ 0.049 & 0.570 $\pm$ 0.133 & 0.706 $\pm$ 0.051 & 0.579 $\pm$ 0.130 \\
 & \textbf{EDGEPRE} & 0.786 $\pm$ 0.093 & \textbf{0.721 $\pm$ 0.175} & \textbf{0.825 $\pm$ 0.108} & \textbf{0.791 $\pm$ 0.168} & \textbf{0.811 $\pm$ 0.091} & \textbf{0.772 $\pm$ 0.153} & \textbf{0.822 $\pm$ 0.112} & \textbf{0.778 $\pm$ 0.188} & \textbf{0.827 $\pm$ 0.108} & \textbf{0.797 $\pm$ 0.161} & \textbf{0.826 $\pm$ 0.109} & \textbf{0.784 $\pm$ 0.167} \\
\midrule
SC3 & BGRL & 0.384 $\pm$ 0.344 & 0.500 $\pm$ 0.000 & 0.384 $\pm$ 0.344 & 0.500 $\pm$ 0.000 & 0.384 $\pm$ 0.344 & 0.500 $\pm$ 0.000 & 0.384 $\pm$ 0.344 & 0.500 $\pm$ 0.000 & 0.433 $\pm$ 0.349 & 0.500 $\pm$ 0.000 & 0.384 $\pm$ 0.344 & 0.500 $\pm$ 0.000 \\
 & GMM & 0.743 $\pm$ 0.252 & 0.674 $\pm$ 0.081 & 0.742 $\pm$ 0.201 & 0.783 $\pm$ 0.129 & 0.668 $\pm$ 0.227 & 0.731 $\pm$ 0.161 & 0.795 $\pm$ 0.107 & 0.748 $\pm$ 0.136 & \textbf{0.864 $\pm$ 0.150} & \textbf{0.890 $\pm$ 0.156} & 0.852 $\pm$ 0.186 & 0.869 $\pm$ 0.126 \\
 & GRACE & 0.866 $\pm$ 0.000 & 0.426 $\pm$ 0.207 & 0.534 $\pm$ 0.469 & 0.446 $\pm$ 0.181 & \textbf{0.866 $\pm$ 0.000} & 0.396 $\pm$ 0.207 & 0.500 $\pm$ 0.337 & 0.441 $\pm$ 0.189 & 0.534 $\pm$ 0.469 & 0.438 $\pm$ 0.186 & 0.866 $\pm$ 0.000 & 0.429 $\pm$ 0.210 \\
 
 & Kmeans & 0.793 $\pm$ 0.090 & 0.636 $\pm$ 0.087 & 0.741 $\pm$ 0.187 & \textbf{0.891 $\pm$ 0.134} & 0.792 $\pm$ 0.223 & 0.774 $\pm$ 0.181 & 0.779 $\pm$ 0.101 & 0.764 $\pm$ 0.164 & 0.754 $\pm$ 0.178 & 0.820 $\pm$ 0.152 & \textbf{0.907 $\pm$ 0.109} & \textbf{0.934 $\pm$ 0.093} \\
 & Spectral & 0.727 $\pm$ 0.168 & 0.592 $\pm$ 0.137 & 0.710 $\pm$ 0.268 & 0.594 $\pm$ 0.189 & 0.684 $\pm$ 0.314 & 0.709 $\pm$ 0.172 & \textbf{0.875 $\pm$ 0.111} & 0.721 $\pm$ 0.207 & 0.677 $\pm$ 0.231 & 0.594 $\pm$ 0.189 & 0.760 $\pm$ 0.148 & 0.569 $\pm$ 0.146 \\
 & \textbf{INFERGAT} & \textbf{0.893 $\pm$ 0.103} & \textbf{0.876 $\pm$ 0.161} & \textbf{0.850 $\pm$ 0.116} & 0.789 $\pm$ 0.147 & 0.765 $\pm$ 0.158 & \textbf{0.813 $\pm$ 0.157} & 0.664 $\pm$ 0.154 & \textbf{0.793 $\pm$ 0.158} & 0.657 $\pm$ 0.199 & 0.860 $\pm$ 0.186 & 0.769 $\pm$ 0.184 & 0.692 $\pm$ 0.391 \\
\midrule
SC4 & BGRL & 0.368 $\pm$ 0.243 & 0.500 $\pm$ 0.000 & 0.368 $\pm$ 0.243 & 0.500 $\pm$ 0.000 & 0.368 $\pm$ 0.243 & 0.500 $\pm$ 0.000 & 0.368 $\pm$ 0.243 & 0.500 $\pm$ 0.000 & 0.374 $\pm$ 0.242 & 0.500 $\pm$ 0.000 & 0.368 $\pm$ 0.243 & 0.500 $\pm$ 0.000 \\
 & GMM & 0.360 $\pm$ 0.116 & 0.292 $\pm$ 0.129 & 0.328 $\pm$ 0.137 & 0.274 $\pm$ 0.149 & 0.335 $\pm$ 0.163 & 0.324 $\pm$ 0.160 & 0.331 $\pm$ 0.141 & 0.309 $\pm$ 0.146 & 0.303 $\pm$ 0.150 & 0.270 $\pm$ 0.157 & 0.299 $\pm$ 0.133 & 0.255 $\pm$ 0.149 \\
 & GRACE & 0.394 $\pm$ 0.244 & 0.536 $\pm$ 0.191 & 0.394 $\pm$ 0.244 & 0.522 $\pm$ 0.188 & 0.393 $\pm$ 0.241 & 0.505 $\pm$ 0.190 & 0.392 $\pm$ 0.244 & 0.510 $\pm$ 0.194 & 0.401 $\pm$ 0.247 & 0.538 $\pm$ 0.190 & 0.395 $\pm$ 0.248 & 0.548 $\pm$ 0.187 \\
 
 & Kmeans & 0.389 $\pm$ 0.097 & 0.272 $\pm$ 0.122 & 0.373 $\pm$ 0.125 & 0.234 $\pm$ 0.129 & 0.359 $\pm$ 0.136 & 0.280 $\pm$ 0.155 & 0.371 $\pm$ 0.121 & 0.269 $\pm$ 0.135 & 0.356 $\pm$ 0.115 & 0.240 $\pm$ 0.144 & 0.344 $\pm$ 0.128 & 0.230 $\pm$ 0.141 \\
 & Spectral & 0.448 $\pm$ 0.136 & 0.448 $\pm$ 0.136 & 0.439 $\pm$ 0.110 & 0.439 $\pm$ 0.110 & 0.465 $\pm$ 0.141 & 0.465 $\pm$ 0.141 & 0.443 $\pm$ 0.118 & 0.443 $\pm$ 0.118 & 0.456 $\pm$ 0.116 & 0.456 $\pm$ 0.116 & 0.435 $\pm$ 0.169 & 0.435 $\pm$ 0.169 \\
 & \textbf{INFERGAT} & \textbf{0.643 $\pm$ 0.194} & \textbf{0.790 $\pm$ 0.140} & \textbf{0.604 $\pm$ 0.232} & \textbf{0.781 $\pm$ 0.132} & \textbf{0.594 $\pm$ 0.201} & \textbf{0.771 $\pm$ 0.134} & \textbf{0.607 $\pm$ 0.179} & \textbf{0.757 $\pm$ 0.148} & \textbf{0.660 $\pm$ 0.185} & \textbf{0.794 $\pm$ 0.145} & \textbf{0.673 $\pm$ 0.199} & \textbf{0.805 $\pm$ 0.152} \\
    \bottomrule
    \end{tabular}
    }
    \caption{F1 and AUC (mean $\pm$ std) across datasets. Best per scenario/dataset in \textbf{bold}.}
    \label{tab:allalgs}
    \end{table*}

Regarding similarity-based metrics, the cosine similarity between connected models is distinctly higher than that between non-connected models, displaying no overlap between the distributions. On the other hand, the curvature divergence and Euclidean similarity show a considerably more significant degree of overlap. Thus, the cosine similarity has been adopted as the attack metric in scenarios where only local models are available (attack scenarios 2 and 4). The complete set of experimental results for all datasets and topologies is in the Appendix.

\subsection{Attack Performance}
After establishing the attack metrics, the proposed topology inference attacks are evaluated on six datasets under four scenarios. As shown in Table~\ref{tab:allalgs}, \textit{EDGEPRE} dominates in SC1–SC2, while \textit{INFERGAT} ranks best in SC3–SC4 across all datasets.

Performance decreases as the attacker’s prior knowledge is reduced from SC1 to SC4. In SC1, \textit{EDGEPRE} attains F1 of at least $\approx$0.85 on every dataset (with high AUC as well). Moving to SC2 yields only a modest drop (typically a few points), indicating that model parameters alone already encode most of the topology signal. The larger degradation appears when supervision is removed: in SC3, \textit{INFERGAT} remains strongest but F1 becomes dataset-dependent (e.g., $\approx$0.89 on CIFAR10 vs. $\approx$0.66–0.77 on MNIST and Malware). Under the weakest prior (SC4), \textit{INFERGAT} still outperforms other unsupervised baselines, achieving F1 around 0.60–0.67 with AUC $\approx$0.76–0.80. Overall, these trends confirm that stronger prior/model knowledge markedly improves topology inference, whereas limited priors and no labels (SC3 and SC4) lead to the main loss in accuracy.

\begin{figure}[t]
    \centering
    \includegraphics[width=0.9\columnwidth]{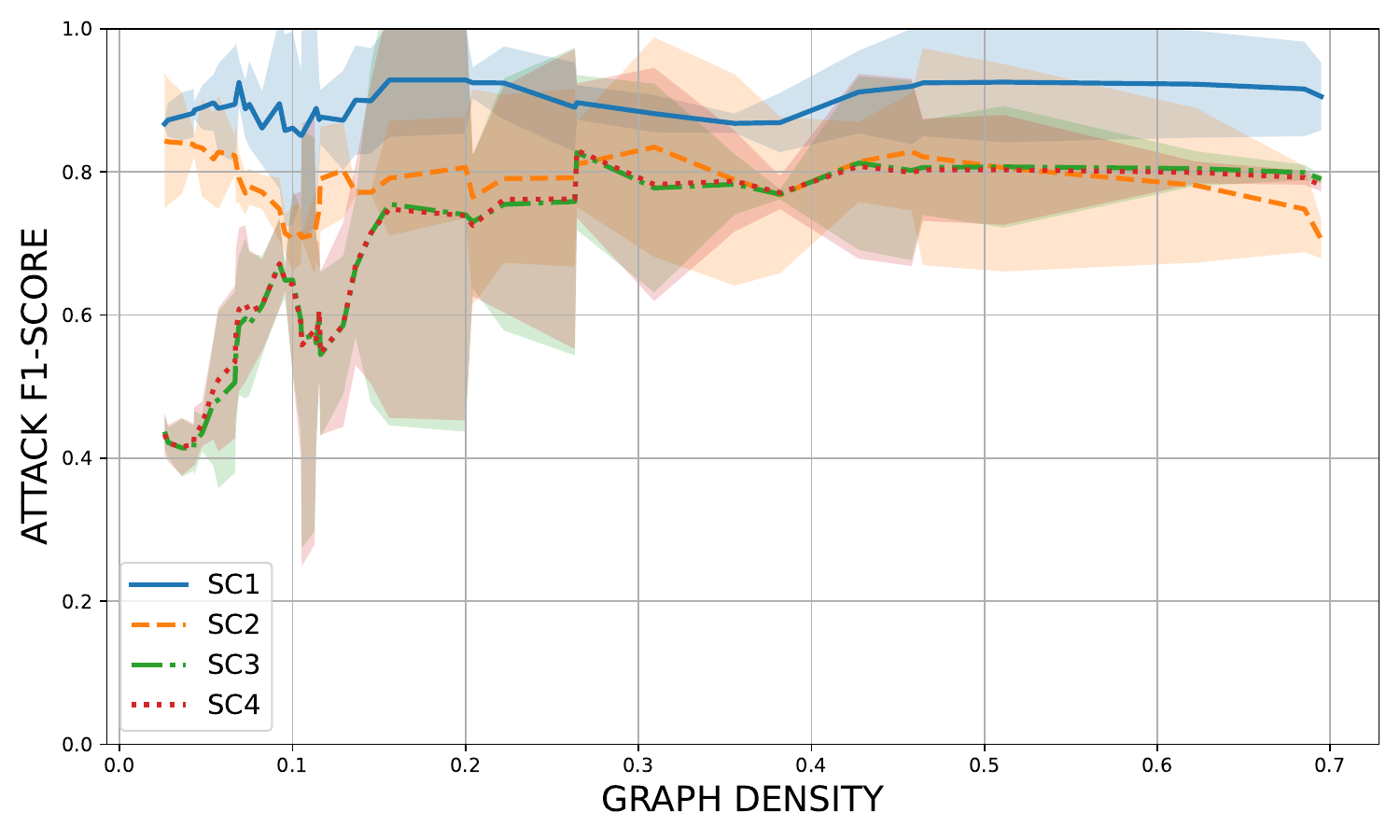}
    \caption{Impact of topology density on the F1-scores achieved by each attack scenario on the CIFAR10 dataset.}
    \label{fig:density}
\end{figure}

\textbf{A. Effect of Topology Density.}
To investigate the influence of topological characteristics on topology inference attacks, \figurename~\ref{fig:density} reports the F1-Scores for four attack scenarios on the CIFAR10 dataset. Topology density, defined as $\frac{2E}{N(N-1)}$, reflects overall connectivity. 

Overall, the four scenarios exhibit relatively stable performance across the full range of densities. For SC3 and SC4, which represent unsupervised attacks, denser topologies provide richer edge information that supports the recovery of the topology, leading to gradually improved performance as connectivity increases. In contrast, SC1 and SC2 achieve comparatively high scores across all densities, reflecting their ability to exploit both sparse and dense settings. The shaded areas further indicate that variability is higher at very low densities, where limited connections introduce uncertainty, while the results become stable once sufficient connectivity is reached.

\begin{figure}[H]
    \centering
    \includegraphics[width=0.9\columnwidth]{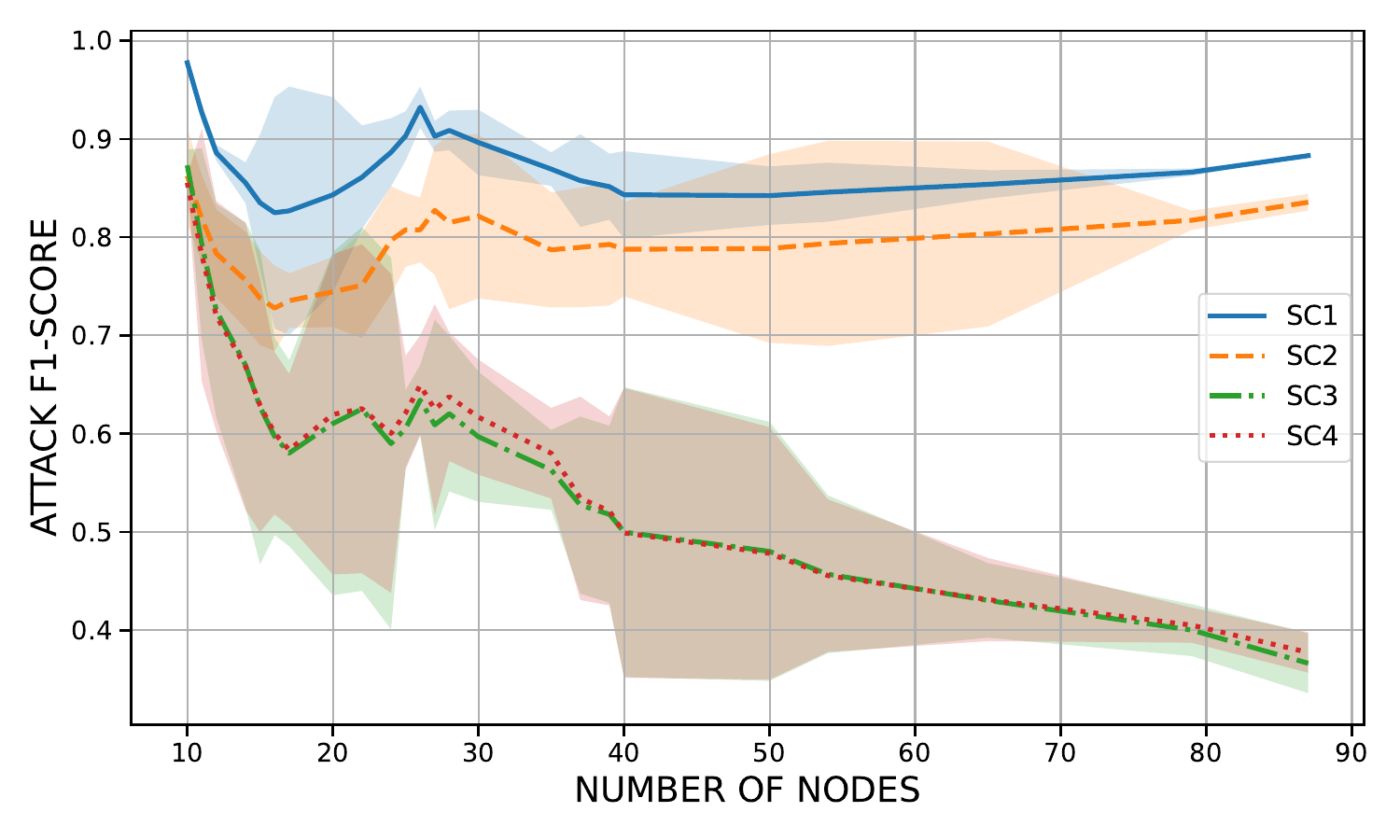}
    \caption{Impact of network size on the F1-scores achieved by each attack scenario on the CIFAR10 dataset.}
    \label{fig:nodes}
\end{figure}

\newpage
\textbf{B. Effect of Network Size.}
To evaluate the scalability of the inference algorithms under different attack scenarios, \figurename~\ref{fig:nodes} illustrates the variation in F1-Scores as the network size increases. SC1 and SC2 remain relatively stable across different numbers of nodes, indicating robustness to changes in network size. In contrast, SC3 and SC4 exhibit a clear decline as the number of nodes grows. This behavior is explained by the fact that larger networks typically correspond to lower densities, which reduce the amount of effective edge information available for unsupervised inference. Consequently, the performance of SC3 and SC4 degrades with increasing network size, whereas the supervised scenarios (SC1 and SC2) maintain more stable effectiveness.

\section{Mitigation}
The effectiveness of the proposed topology inference attacks indicates that topology in DFL systems can be discerned solely by analyzing model behaviors. This observation highlights the security and privacy concerns of DFL systems. This section proposes defense strategies to mitigate these vulnerabilities within DFL environments.

\begin{figure}[b]
    \centering
    \includegraphics[width=1\columnwidth]{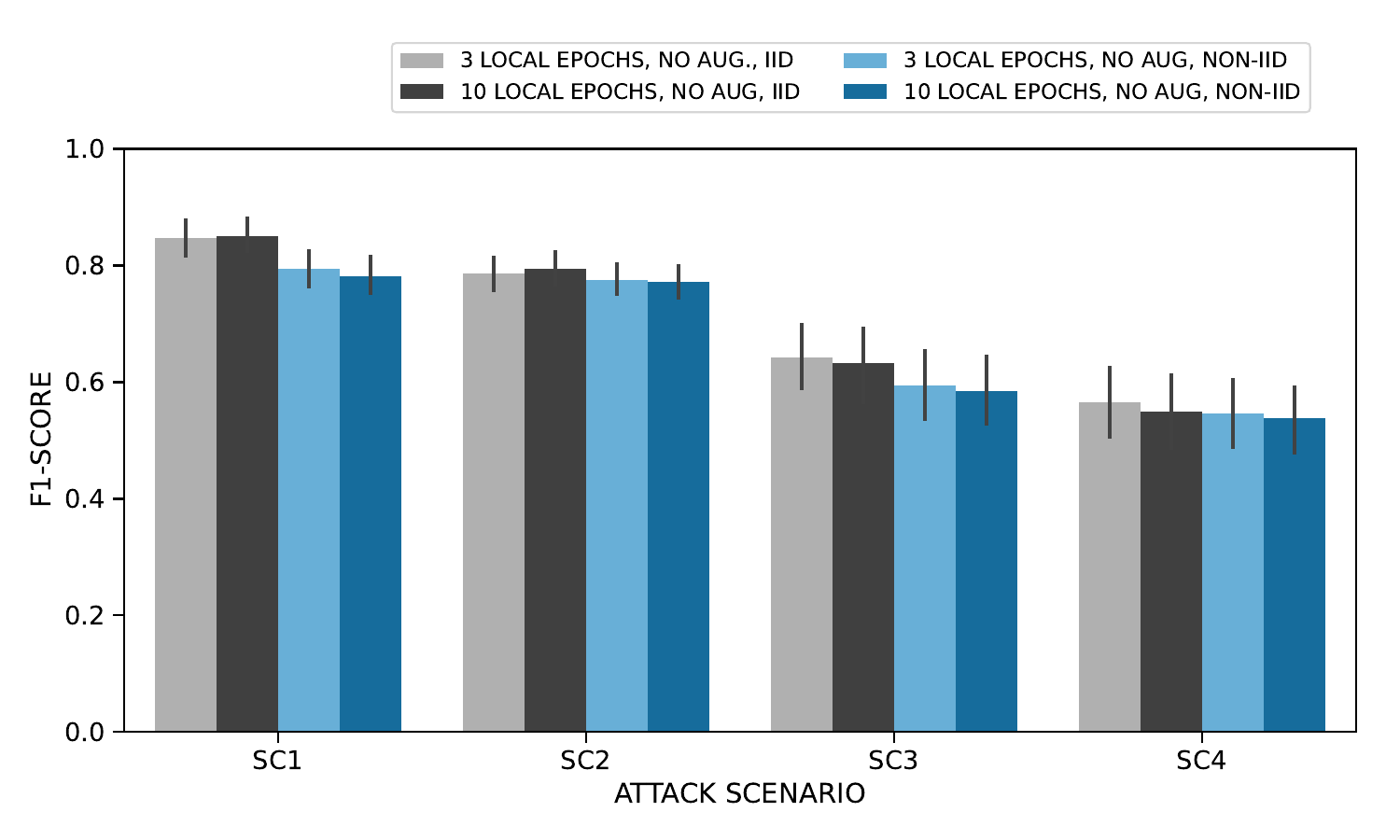}
    \caption{Impact of \textbf{local training epochs} and\textbf{ data heterogeneity} on the F1-Score of topology inference attacks in the CIFAR10 dataset.}
    \label{fig:mitigation}
\end{figure}

\subsection{Reduce Overfitting}

As illustrated in \figurename~\ref{fig:mitigation}, each bar represents a specific training configuration: gray bars denote IID data distributions without augmentation and blue bars correspond to non-IID data distributions (Dirichlet $\alpha = 0.1$).

Consistent with findings in prior inference attack literature, the success of topology inference attacks is closely tied to the degree of overfitting in the target model. To validate this, experiments were conducted on the CIFAR10 dataset under four attack scenarios. By increasing the number of local training epochs from 3 to 10, the level of overfitting was elevated. Results show that, across all examined topologies, this change consistently improves the F1-Scores of attacks. This observation confirms that overfitting exacerbates the risk of topological information leakage in DFL systems.

\subsection{Data Heterogeneity}
Furthermore, data distribution among clients plays a critical role in attack effectiveness. Non-IID data were simulated using a Dirichlet distribution with $\alpha = 0.1$ to assess this. The results, represented by the blue bars, indicate a noticeable decline in attack performance across all scenarios, with the effect being most prominent in scenarios 3 and 4. This suggests that increased data heterogeneity introduces higher model divergence, which in turn hampers the inference model’s ability to exploit overfitted representations for edge prediction.

\subsection{Differential Privacy}
Differential Privacy (DP) is a common mitigation for privacy in FL~\cite{10251949}. 
DP is integrated into DFL parameter sharing and its effect on topology inference is assessed. 
On CIFAR10 (ER\_0.5) under SC1, attack F1 drops from \(0.889 \rightarrow 0.687\) (10 nodes) and \(0.824 \rightarrow 0.637\) (20 nodes), approximately \(23\%\) reductions, indicating effective mitigation. 

Overall, these findings imply that reducing overfitting through fewer local training epochs and increasing data heterogeneity are promising directions for mitigating topology inference risks in DFL. 

\section{Conclusion and Future Work}
This work introduces a novel topology inference attack against DFL, exposing critical vulnerabilities related to privacy and information leakage. By analyzing local models, attackers can accurately infer the overlay topology, one of DFL's most sensitive assets, highlighting the system's susceptibility to privacy breaches. The study explores various attack scenarios and develops tailored metrics, models, and algorithms, with experiments confirming the feasibility and effectiveness of these attacks.  Furthermore, network size, density, model overfitting, and data heterogeneity significantly influence attack success. Mitigation strategies, such as increasing data heterogeneity, can enhance model generalization and reduce the risk of data leakage.

The present work focuses on attack scenarios 1 through 4 and does not yet address the most challenging attack scenario 5. Future research plans include exploring feasible strategies for Scenario 5 and extending the evaluations to a wider range of datasets and topologies. Besides, more attack metrics and strategies will be developed to improve attack effectiveness.

\section*{Acknowledgment}
This work has been partially supported by \textit{(a)} the Swiss Federal Office for Defense Procurement (armasuisse) with the CyberDFL project (CYD-C-2020003) and \textit{(b)} the University of Zürich UZH.

%%%%%%%%%%%%%%%%%%%%%%%%%%%%%%%%%%%%%%%%%%%%%%%%%%%%%%%%%%%%%%%%%%%%%%%%

%%% Use this command to include your bibliography file.

\bibliography{mybibfile}
% \balance

\onecolumn
% \tableofcontents
\appendix
\newpage

\section{appendix}

%%% Use this command to specify the title of your paper.

\subsection{Topology Information}
This work utilizes two categories of topologies: synthetic and real-world. The synthetic topologies include ring, star, and three Erdős-Rényi (ER) graphs with connection probabilities \( p = \)0.3, 0.5, and 0.7. The real-world topologies consist of 27 network topologies collected from the two public network repositories: SNDlib\footnote{Available in: https://sndlib.put.poznan.pl/home.action} and DEFO\footnote{Available in: https://sites.uclouvain.be/defo/}, which reflect realistic communication and infrastructure networks.

To characterize the structure of different graphs, this work reports the following topological statistics:

\begin{itemize}
    \item \textbf{Number of Nodes}: The total number of distinct vertices in the graph, denoted as $N$.
    
    \item \textbf{Number of Edges}: The total number of edges connecting node pairs, denoted as $E$. 
    
    \item \textbf{Average Degree}: The mean number of edges connected to each node, calculated as $\frac{2E}{N}$.
    
    \item \textbf{Density}: The ratio of the number of actual edges to the maximum possible number of edges, given by $\frac{2E}{N(N-1)}$. This metric reflects the graph's overall connectivity or sparsity.
\end{itemize}

\subsubsection{Synthetic Topology Characterize Statistics}
To evaluate the robustness and generalizability of the proposed method under various topology conditions, this work considers a set of synthetic topologies ranging from sparse to dense:

\begin{itemize}
    \item \textbf{STAR graphs} represent extremely sparse topologies, where most nodes are only connected to a single central hub. These graphs have low average degree and density, reflecting minimal connectivity.
    
    \item \textbf{RING graphs} are sparse structures with uniform and low node degrees. While slightly denser than STAR graphs, they remain sparse and exhibit limited path diversity.
    
    \item \textbf{ER graphs} with varying edge probabilities ($p=0.3$, $0.5$, and $0.7$) capture a continuum of connectivity levels. As $p$ increases, the average degree and edge density grows, gradually transitioning the graph from a sparse to a dense structure.
\end{itemize}

This selection enables controlled evaluation across different topological regimes, ensuring that performance is not biased toward a specific graph sparsity or structure. The characterize statistics of synthetic topologies are shown in Table~\ref{tab:graph_stats}

\begin{table}[htbp]
\centering
\caption{Topological Statistics of Different Synthetic Topology}
\begin{tabular}{lcccc}
\toprule
\textbf{Topology} & \textbf{Number of Nodes} & \textbf{Number of Edges} & \textbf{Average Degree} & \textbf{Density} \\
\midrule
ER($p=0.3$) & 10 & 16 & 3.20 & 0.36 \\
ER($p=0.3$) & 30 & 115 & 7.67 & 0.26 \\
ER($p=0.3$) & 20 & 50 & 5.00 & 0.26 \\
ER($p=0.5$) & 10 & 23 & 4.60 & 0.51 \\
ER($p=0.5$) & 30 & 202 & 13.47 & 0.46 \\
ER($p=0.5$) & 20 & 87 & 8.70 & 0.46 \\
ER($p=0.7$) & 20 & 132 & 13.20 & 0.69 \\
ER($p=0.7$) & 30 & 298 & 19.87 & 0.69 \\
ER($p=0.7$) & 10 & 28 & 5.60 & 0.62 \\
RING & 10 & 10 & 2.00 & 0.22 \\
RING & 20 & 20 & 2.00 & 0.11 \\
RING & 30 & 30 & 2.00 & 0.07 \\
STAR & 10 & 9 & 1.80 & 0.20 \\
STAR & 20 & 19 & 1.90 & 0.10 \\
STAR & 30 & 29 & 1.93 & 0.07 \\
\bottomrule
\end{tabular}
\label{tab:graph_stats}
\end{table}

\subsubsection{Real-world Topology Characterize Statistics}

To evaluate the generalizability and robustness of the proposed approach across diverse topological structures, experiments were conducted on a wide range of real-world network topologies. These networks vary significantly in size, ranging from small-scale graphs with only 10 nodes to large-scale structures with up to 87 nodes. The topologies are sourced from realistic communication infrastructures, including national research and education networks (NRENs) and Internet service provider (ISP) backbone networks, which reflect actual deployment scenarios.

A statistical overview of these networks is provided in Table~\ref{tab:realworld_topologies}. Compared to synthetic graphs, real-world networks exhibit higher structural variability in terms of node degree, connectivity, and scale. Despite this diversity, they are generally characterized by sparse connectivity, as indicated by low edge densities and moderate average degrees. This sparsity is a common property of real communication networks and introduces specific challenges to graph data.

\begin{table}[htbp]
\centering
\caption{Topological statistics of real-world and synthetic networks}
\begin{tabular}{lcccc}
\toprule
\textbf{Topology} & \textbf{Number of Nodes} & \textbf{Number of Edges} & \textbf{Average Degree} & \textbf{Density} \\
\midrule
ABILENE & 12 & 15 & 2.50 & 0.11 \\
ATLANTA & 15 & 22 & 2.93 & 0.10 \\
COST266 & 37 & 57 & 3.08 & 0.04 \\
DFN-GWIN & 11 & 47 & 8.55 & 0.43 \\
DI-YUAN & 11 & 42 & 7.64 & 0.38 \\
FRANCE & 25 & 45 & 3.60 & 0.08 \\
GEANT & 22 & 72 & 6.55 & 0.16 \\
GERMANY50 & 50 & 88 & 3.52 & 0.04 \\
GIUL39 & 39 & 172 & 8.82 & 0.12 \\
INDIA35 & 35 & 80 & 4.57 & 0.07 \\
JANOS-US & 26 & 84 & 6.46 & 0.13 \\
JANOS-US-CA & 39 & 122 & 6.26 & 0.08 \\
NEWYORK & 16 & 49 & 6.13 & 0.20 \\
NOBEL-EU & 28 & 41 & 2.93 & 0.05 \\
NOBEL-GERMANY & 17 & 26 & 3.06 & 0.10 \\
NOBEL-US & 14 & 21 & 3.00 & 0.12 \\
NORWAY & 27 & 51 & 3.78 & 0.07 \\
PDH & 11 & 34 & 6.18 & 0.31 \\
PIORO40 & 40 & 89 & 4.45 & 0.06 \\
POLSKA & 12 & 18 & 3.00 & 0.14 \\
RF1755 & 87 & 322 & 7.40 & 0.04 \\
RF3967 & 79 & 294 & 7.44 & 0.05 \\
SUN & 27 & 102 & 7.56 & 0.15 \\
SYNTH50 & 50 & 276 & 11.04 & 0.11 \\
TA1 & 24 & 51 & 4.25 & 0.09 \\
TA2 & 65 & 108 & 3.32 & 0.03 \\
ZIB54 & 54 & 80 & 2.96 & 0.03 \\
\bottomrule
\end{tabular}
\label{tab:realworld_topologies}
\end{table}

\subsection{Attack Metrics Distribution}
\begin{figure}[b]
    \centering
    \includegraphics[width=1\columnwidth]{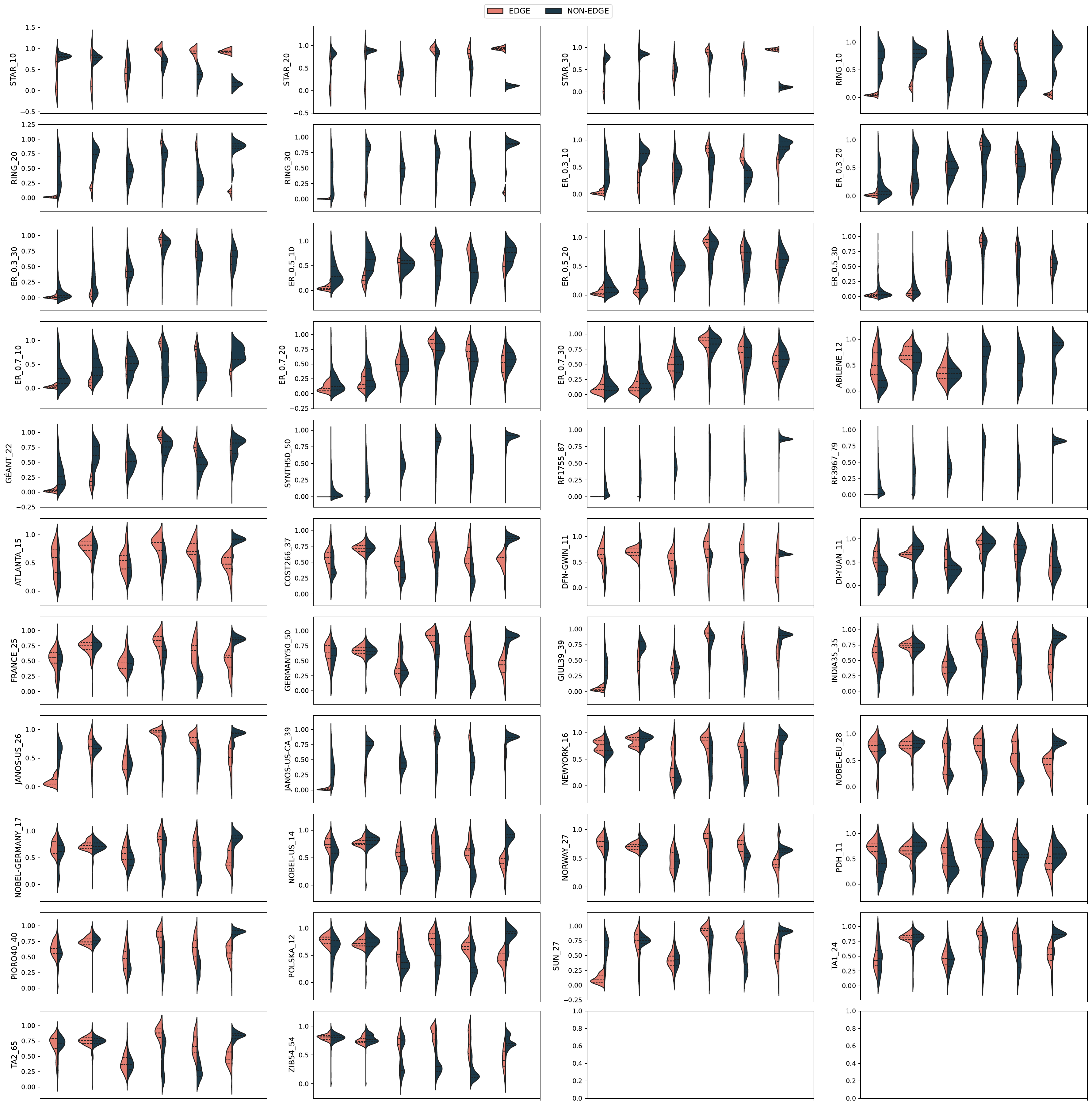}
    \caption{Normalized distributions of attack metrics in various topology DFL systems trained on the \textbf{CIFAR10} dataset. The “edge group” corresponds to metrics calculated between directly connected nodes, while the “non-edge group” corresponds to metrics calculated between nodes with no direct connection. In each subfigure, the attack metrics are respectively: Relative Loss, Relative Entropy, Relative Sensitivity, Cosine Similarity, Euclidean Similarity, and Curvature Divergence.}
    \label{fig:cifar10}
\end{figure}

\begin{figure}[b]
    \centering
    \includegraphics[width=1\columnwidth]{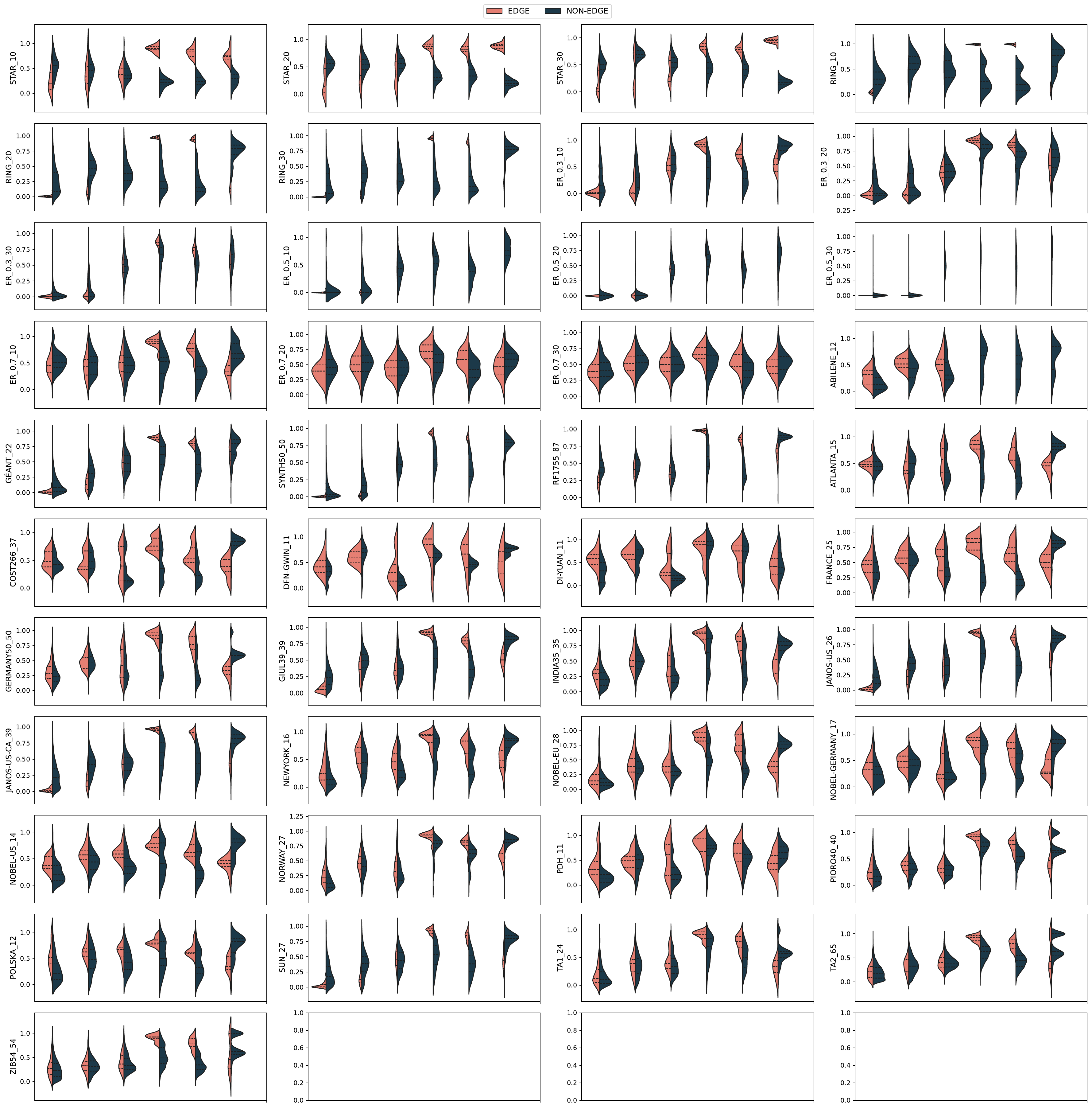}
    \caption{Normalized distributions of attack metrics in various topology DFL systems trained on the \textbf{MNIST} dataset. The “edge group” corresponds to metrics calculated between directly connected nodes, while the “non-edge group” corresponds to metrics calculated between nodes with no direct connection. In each subfigure, the attack metrics are respectively: Relative Loss, Relative Entropy, Relative Sensitivity, Cosine Similarity, Euclidean Similarity, and Curvature Divergence.}
    \label{fig:mnist}
\end{figure}

\begin{figure}[b]
    \centering
    \includegraphics[width=1\columnwidth]{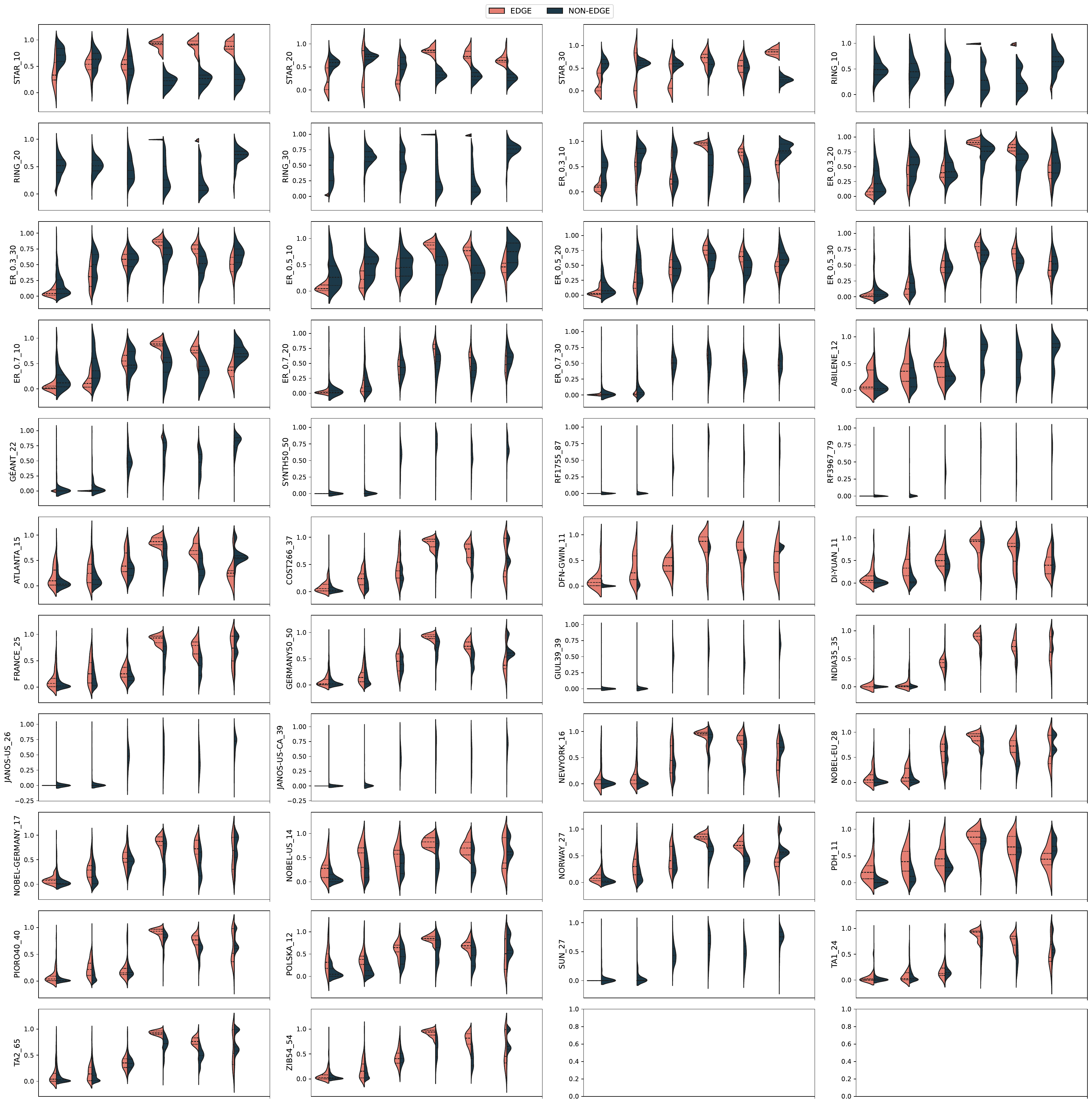}
    \caption{Normalized distributions of attack metrics in various topology DFL systems trained on the \textbf{FMNIST} dataset. The “edge group” corresponds to metrics calculated between directly connected nodes, while the “non-edge group” corresponds to metrics calculated between nodes with no direct connection. In each subfigure, the attack metrics are respectively: Relative Loss, Relative Entropy, Relative Sensitivity, Cosine Similarity, Euclidean Similarity, and Curvature Divergence.}
    \label{fig:fmnist}
\end{figure}

\begin{figure}[b]
    \centering
    \includegraphics[width=1\columnwidth]{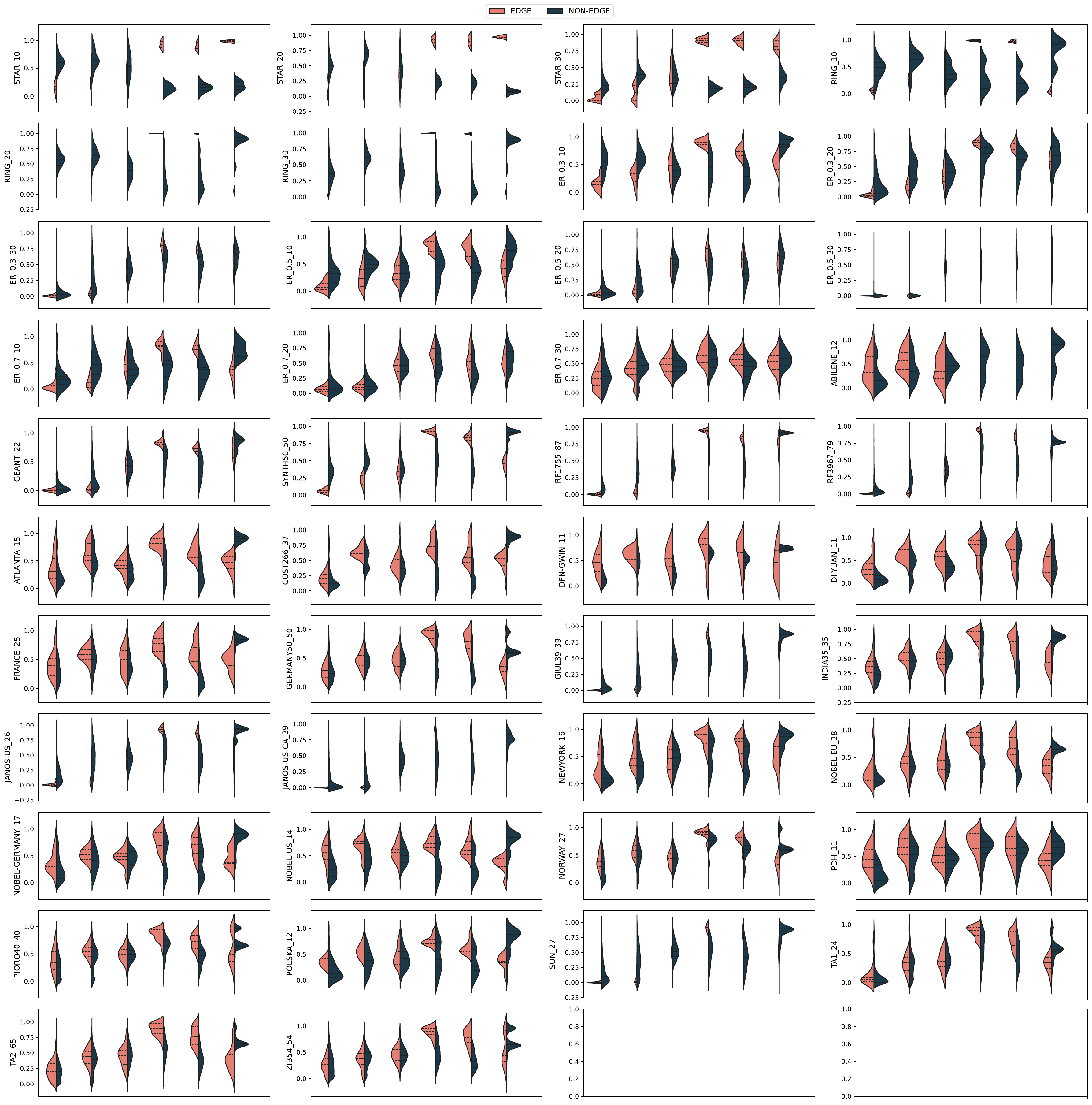}
    \caption{Normalized distributions of attack metrics in various topology DFL systems trained on the \textbf{SVHN} dataset. The “edge group” corresponds to metrics calculated between directly connected nodes, while the “non-edge group” corresponds to metrics calculated between nodes with no direct connection. In each subfigure, the attack metrics are respectively: Relative Loss, Relative Entropy, Relative Sensitivity, Cosine Similarity, Euclidean Similarity, and Curvature Divergence.}
    \label{fig:svhn}
\end{figure}

\begin{figure}[b]
    \centering
    \includegraphics[width=1\columnwidth]{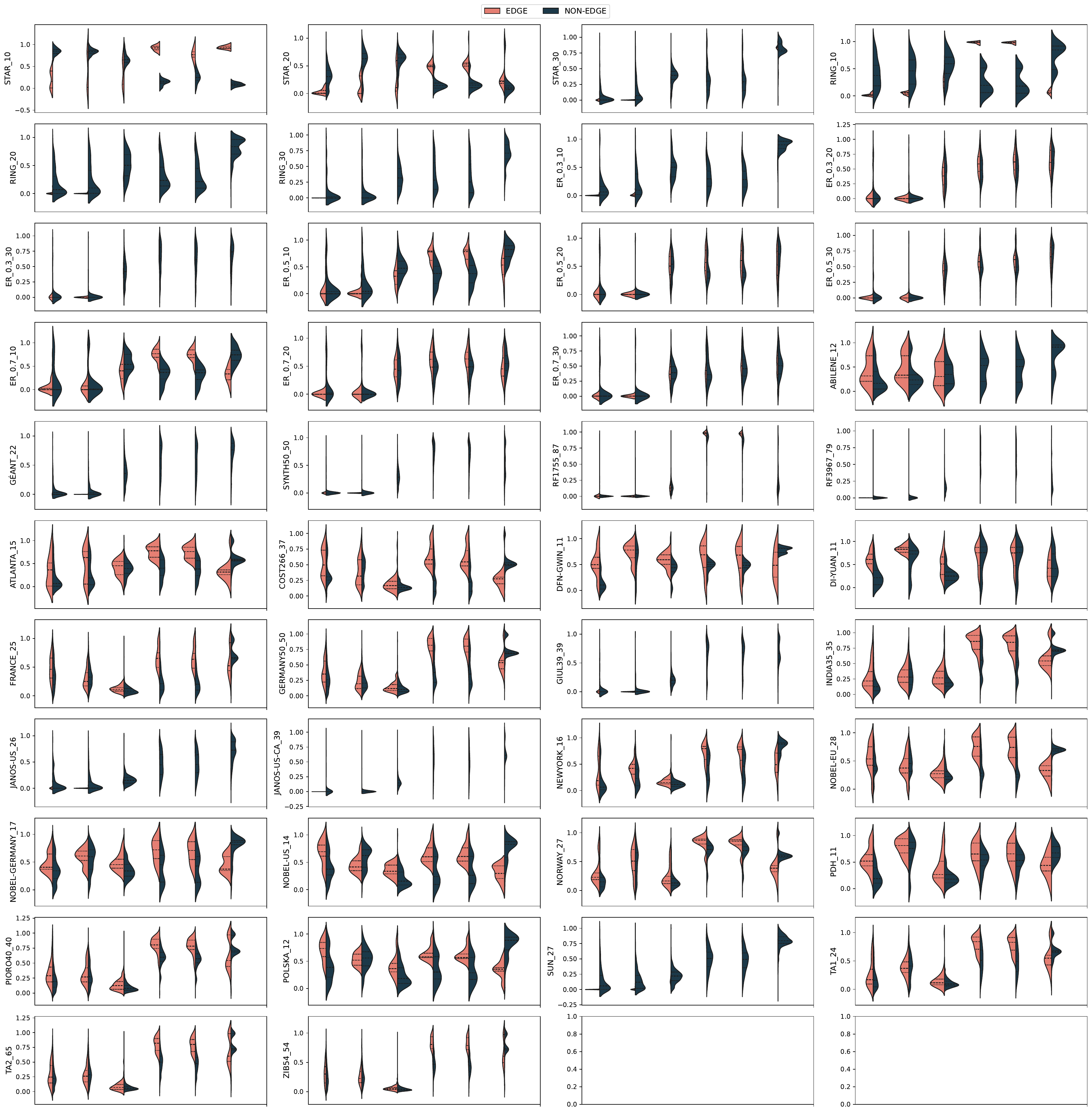}
    \caption{Normalized distributions of attack metrics in various topology DFL systems trained on the \textbf{ImageNet10} dataset. The “edge group” corresponds to metrics calculated between directly connected nodes, while the “non-edge group” corresponds to metrics calculated between nodes with no direct connection. In each subfigure, the attack metrics are respectively: Relative Loss, Relative Entropy, Relative Sensitivity, Cosine Similarity, Euclidean Similarity, and Curvature Divergence.}
    \label{fig:imagenet}
\end{figure}

\begin{figure}[b]
    \centering
    \includegraphics[width=1\columnwidth]{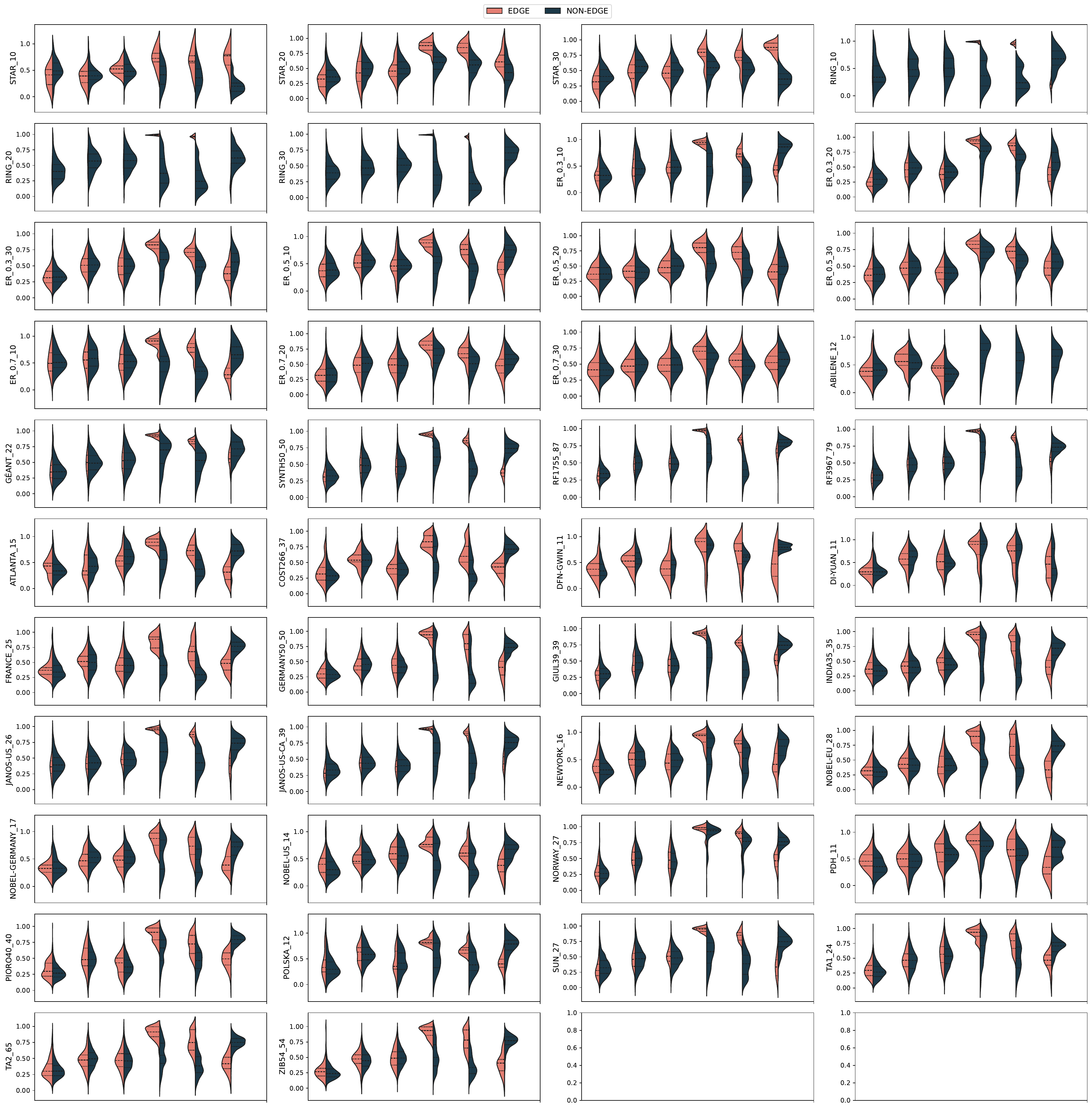}
    \caption{Normalized distributions of attack metrics in various topology DFL systems trained on the \textbf{Malware} dataset. The “edge group” corresponds to metrics calculated between directly connected nodes, while the “non-edge group” corresponds to metrics calculated between nodes with no direct connection. In each subfigure, the attack metrics are respectively: Relative Loss, Relative Entropy, Relative Sensitivity, Cosine Similarity, Euclidean Similarity, and Curvature Divergence.}
    \label{fig:malware}
\end{figure}

This experiment comprehensively demonstrates the distribution of attack metrics in the edge and non-edge groups across CIFAR10 (\figurename~\ref{fig:cifar10}), MNIST(\figurename~\ref{fig:mnist}), FMNIST(\figurename~\ref{fig:fmnist}), SVHN(\figurename~\ref{fig:svhn}), ImageNet10 (\figurename~\ref{fig:imagenet}), and Malware (\figurename~\ref{fig:malware})  datasets in various topologies. It can be observed that, in most cases, the distribution of relative loss is more concentrated than that of relative entropy and sensitivity. Additionally, relative loss exhibits less overlap between the edge group and the non-edge group. Therefore, this work selects relative loss as the attack metric when the attacker has access to both local data and the local model, \textit{i.e.,} attack scenarios 1 and 3.

Regarding similarity-based metrics, the cosine similarity between connected models is distinctly higher than that between non-connected models, displaying no overlap between the distributions. On the other hand, the curvature divergence and Euclidean similarity show a considerably more significant degree of overlap. Thus, the cosine similarity has been adopted as the attack metric in scenarios where only local models are available (attack scenarios 2 and 4).

\end{document}